\documentclass[10pt,logo,copyright]{giga-report}
\usepackage[authoryear,sort&compress,round]{natbib}

\usepackage[utf8]{inputenc} %
\usepackage[T1]{fontenc}    %

\usepackage{parskip}        %
\usepackage{url}            %
\usepackage{booktabs}       %
\usepackage{amsfonts}       %
\usepackage{nicefrac}       %
\usepackage{microtype}      %
\usepackage{xcolor}         %
\usepackage[dvipsnames]{xcolor} %
\usepackage{graphicx}
\usepackage{animate}        %
\usepackage{subcaption}
\usepackage{tabularx}
\usepackage{makecell}
\usepackage{adjustbox}
\usepackage{setspace}
\usepackage{todonotes}
\usepackage{colortbl}
\usepackage{wrapfig}

\newcolumntype{M}[1]{>{\centering\arraybackslash}m{#1}}
\usepackage{float}
\usepackage{tikz}
\usetikzlibrary{positioning,shapes,arrows}
\usepackage{amsmath,amsfonts,bm, bbm,leftindex}
\usepackage{multirow}
\usepackage{comment}
\usepackage{gensymb}
\usepackage{lipsum}
\usetikzlibrary{arrows.meta, positioning, fit}
\usepackage[para]{threeparttable}
\usepackage{tikz}
\usetikzlibrary{tikzmark}

\def\eqref#1{equation~\ref{#1}}

\def\1{\bm{1}}

\DeclareMathAlphabet{\mathsfit}{\encodingdefault}{\sfdefault}{m}{sl}
\SetMathAlphabet{\mathsfit}{bold}{\encodingdefault}{\sfdefault}{bx}{n}

\makeatletter
\let\save@mathaccent\mathaccent
\newcommand*\if@single[3]{%
  \setbox0\hbox{${\mathaccent"0362{#1}}^H$}%
  \setbox2\hbox{${\mathaccent"0362{\kern0pt#1}}^H$}%
  \ifdim\ht0=\ht2 #3\else #2\fi
  }
\newcommand*\rel@kern[1]{\kern#1\dimexpr\macc@kerna}
\newcommand*\widebar[1]{\@ifnextchar^{{\wide@bar{#1}{0}}}{\wide@bar{#1}{1}}}
\newcommand*\wide@bar[2]{\if@single{#1}{\wide@bar@{#1}{#2}{1}}{\wide@bar@{#1}{#2}{2}}}
\newcommand*\wide@bar@[3]{%
  \begingroup
  \def\mathaccent##1##2{%
    \let\mathaccent\save@mathaccent
    \if#32 \let\macc@nucleus\first@char \fi
    \setbox\z@\hbox{$\macc@style{\macc@nucleus}_{}$}%
    \setbox\tw@\hbox{$\macc@style{\macc@nucleus}{}_{}$}%
    \dimen@\wd\tw@
    \advance\dimen@-\wd\z@
    \divide\dimen@ 3
    \@tempdima\wd\tw@
    \advance\@tempdima-\scriptspace
    \divide\@tempdima 10
    \advance\dimen@-\@tempdima
    \ifdim\dimen@>\z@ \dimen@0pt\fi
    \rel@kern{0.6}\kern-\dimen@
    \if#31
      \overline{\rel@kern{-0.6}\kern\dimen@\macc@nucleus\rel@kern{0.4}\kern\dimen@}%
      \advance\dimen@0.4\dimexpr\macc@kerna
      \let\final@kern#2%
      \ifdim\dimen@<\z@ \let\final@kern1\fi
      \if\final@kern1 \kern-\dimen@\fi
    \else
      \overline{\rel@kern{-0.6}\kern\dimen@#1}%
    \fi
  }%
  \macc@depth\@ne
  \let\math@bgroup\@empty \let\math@egroup\macc@set@skewchar
  \mathsurround\z@ \frozen@everymath{\mathgroup\macc@group\relax}%
  \macc@set@skewchar\relax
  \let\mathaccentV\macc@nested@a
  \if#31
    \macc@nested@a\relax111{#1}%
  \else
    \def\gobble@till@marker##1\endmarker{}%
    \futurelet\first@char\gobble@till@marker#1\endmarker
    \ifcat\noexpand\first@char A\else
      \def\first@char{}%
    \fi
    \macc@nested@a\relax111{\first@char}%
  \fi
  \endgroup
}
\makeatother

\definecolor{darkred}{rgb}{0.7, 0.0, 0.0}

\usepackage{amsmath} 

\usepackage[table]{xcolor}

\usepackage{pifont}
\usepackage{graphicx}

\usepackage[nameinlink]{cleveref}
\crefname{equation}{Eq.}{Eqs.}
\crefname{figure}{Fig.}{Figs.}
\crefname{section}{Sec.}{Sec.}
\crefname{appendix}{App.}{App.}
\crefname{table}{Tab.}{Tabs.}
\crefname{algorithm}{Algo}{Algo}
\crefname{thm}{Thm}{Thm}
\Crefname{thm}{Thm}{Thm}
\crefname{prop}{Prop}{Prop}

\usepackage{graphicx}
\usepackage{booktabs}

\usepackage{caption}
\usepackage{url}
\usepackage{graphicx}

\usepackage{subcaption}
\usepackage{booktabs} 

\usepackage{multirow}
\usepackage{makecell}
\definecolor{lightgrey}{gray}{0.9}
\usepackage{xcolor}
\usepackage{pifont}
\usepackage{amssymb}
\usepackage{bbding}
\usepackage{xcolor,colortbl}
\definecolor{minetable1colorx}{rgb}{0.75, 0.75, 0.75}
\newcommand{\mineyes}{{\scriptsize \CheckmarkBold}}
\newcommand{\mineno}{{\scriptsize \textcolor{minetable1colorx}{\XSolidBrush}}}

\newcommand{\crefnames}[3]{%
  \@for\next:=#1\do{%
    \expandafter\crefname\expandafter{\next}{#2}{#3}%
  }%
}

\title{DriveDreamer-Policy: A Geometry-Grounded World–Action Model for Unified Generation and Planning}

\author{
\vspace{0.3in}
\normalfont

{\large
\centerline{
\textbf{Yang Zhou\textsuperscript{1,2} \quad Xiaofeng Wang\textsuperscript{1} \quad Hao Shao\textsuperscript{3} \quad Letian Wang\textsuperscript{2}}
}

\vspace{0.1in}

\centerline{
\textbf{Guosheng Zhao\textsuperscript{1} \quad Jiangnan Shao\textsuperscript{1} \quad Jiagang Zhu\textsuperscript{1} \quad Tingdong Yu\textsuperscript{1}}
}

\vspace{0.1in}

\centerline{
\textbf{Zheng Zhu\textsuperscript{1,*} \quad Guan Huang\textsuperscript{1} \quad Steven L. Waslander\textsuperscript{2}}
}
}

\vspace{0.1in}
\quad \\

{\large
\centerline{
\textsuperscript{1}{\ GigaAI} \quad
\textsuperscript{2}{\ University of Toronto} \quad
\textsuperscript{3}{\ CUHK MMLab} \quad
\textsuperscript{*}{\ Corresponding Author}}
}

\vspace{0.1in}

{\large
\centerline{
Project Website: \href{https://drivedreamer-policy.github.io/}{\textcolor{magenta}{https://drivedreamer-policy.github.io/}}
}
}
}

\begin{document}
\maketitle


\begin{abstract}

Recently, world–action models (WAM) have emerged to bridge vision–language–action (VLA) models and world models, unifying their reasoning and instruction-following capabilities and spatio-temporal world modeling.
However, existing WAM approaches often focus on modeling 2D appearance or latent representations, with limited geometric grounding-an essential element for embodied systems operating in the physical world.
We present DriveDreamer-Policy, a unified driving world–action model that integrates depth generation, future video generation, and motion planning within a single modular architecture. The model employs a large language model to process language instructions, multi-view images, and actions, followed by three lightweight generators that produce depth, future video, and actions.
By learning a geometry-aware world representation and using it to guide both future prediction and planning within a unified framework, the proposed model produces more coherent imagined futures and more informed driving actions, while maintaining modularity and controllable latency.
Experiments on the Navsim v1 and v2 benchmarks demonstrate that DriveDreamer-Policy achieves strong performance on both closed-loop planning and world generation tasks. In particular, our model reaches 89.2 PDMS on Navsim v1 and 88.7 EPDMS on Navsim v2, outperforming existing world-model-based approaches while producing higher-quality future video and depth predictions. Ablation studies further show that explicit depth learning provides complementary benefits to video imagination and improves planning robustness.
\end{abstract}

\abscontent
\section{Introduction}

Autonomous driving systems have undergone a paradigm shift from handcrafted modular stacks~\citep{stanley} to end-to-end learning systems~\citep{bojarski2016endend,hu2023uniad}, and more recently to vision-language-action (VLA) models built on large language models (LLMs) that offer richer common-sense knowledge, instruction following, and reasoning capabilities \citep{zhou2025opendrivevla,kirby2026drivingregisters,zhou2025autovla,li2025drivevlaw0}. 
Despite these advances, most VLA planners primarily optimize action outputs and do not explicitly model how the future world may evolve under alternative actions, which limits interpretability and can reduce reliability in rare or safety-critical situations where forward-looking reasoning about occlusions and hidden hazards is essential. In parallel, driving world models have emerged as a powerful direction for learning spatio-temporal traffic dynamics directly from large-scale sensor logs \citep{gao2024vista,russell2025gaia2,hassan2024gem,mousakhan2025orbis,li2025omninwm,bartoccioni2025vavimvavam,zhou2026drivinggen,zhao2026unidrivedreamer,ye2026gigaworld}. 
By forecasting future observations (\textit{e.g.}, videos, BEV features, or other state representations), they enable scalable simulation, rare-event evaluation, and controllable synthetic data generation. 
Motivated by the complementary strengths of these two lines, recent world-action models aim to unify future world generation with motion prediction or planning \citep{zhang2025epona,xia2025drivelaw,lu2025uniugp,zhao2025pwm}, thereby bridging imagination and decision-making in a single framework.

However, in many existing world-action models, the world component is still implemented as image/video prediction or latent rollouts without explicit geometric grounding, and the benefits of imagined futures for action prediction can be constrained by representation mismatch or tightly coupled module designs.
As a result, the generated world may be visually plausible but not maximally informative for planning, and the planner may not consistently benefit from structured safety cues such as geometric layout and free-space constraints.


\begin{figure}[t]
    \centering
    \includegraphics[width=0.99\linewidth]
    {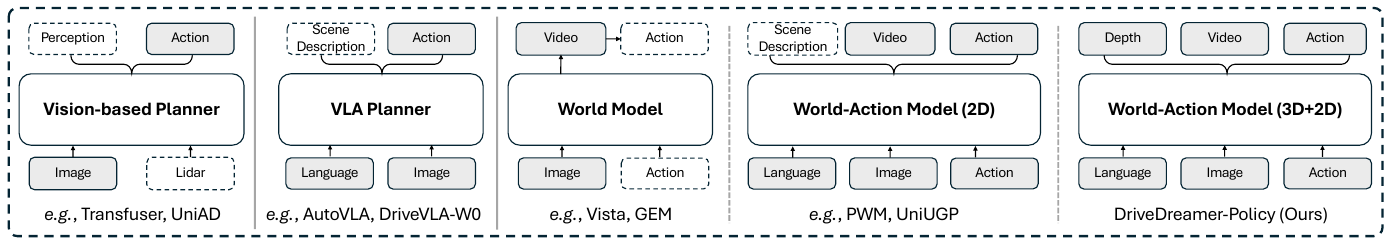}
    \captionsetup{justification=justified, singlelinecheck=false}
    \caption{Comparison of our DriveDreamer-Policy with existing models. Items with dashed lines are optional.
    Vision-based and VLA planners directly map observations (and optional inputs) to actions without explicitly predicting the future world. World models generate future observations but often rely on external action signals. Recent world–action models unify future world generation and planning, but typically operate on image/video representations. DriveDreamer-Policy extends this line by explicitly generating depth alongside video and actions, enabling geometry-grounded imagination and planning within a unified model.
    }
    \label{fig:motiv}
\end{figure}


A key observation motivating this work (Fig.~\ref{fig:motiv}) is that autonomous driving is fundamentally a 4D physical process: 3D geometry evolves over time. Consequently, an actionable world model should not only synthesize appearance, but also preserve geometric structure that is essential for occlusion reasoning, distance estimation, and physically consistent motion. Depth-centric modeling is particularly attractive here: depth is compact, directly tied to geometry, and can serve as an explicit scaffold that constrains future image/video generation and informs planning decisions. Moreover, recent progress in depth foundation models \citep{yang2024depthv2,lin2025depth3,piccinelli2024unidepth,Piccinelli_2026,xu2025pixelperfectdepth} suggests that high-fidelity depth can be produced off-the-shelf, without collecting extra data or training a depth estimator from scratch.
These developments suggest an opportunity to more effectively drive world-action models: explicitly generating depth representations and studying how this benefits both future video generation and motion planning within a unified architecture.

To this end, we propose \textbf{DriveDreamer-Policy}, a unified driving world-action model that jointly generates 1) a depth-based 3D geometric representation of the current scene, 2) action-conditioned future videos, and 3) future trajectories for planning. The system is built on a large language model for perception and reasoning, producing a compact set of world embeddings and action embeddings. These embeddings serve as conditions for the multimodal generator: a pixel-space depth generator, a 
latent-space video generator, and an action generator. Importantly, we impose a structured causal attention mask across query groups in a {depth$\rightarrow$video$\rightarrow$action} manner: video queries may consume depth context, and action queries may consume both depth and video context. This yields a simple, single-pass information flow, while enabling video imagination to benefit from 3D understanding and allow planning to leverage both 3D structure and predicted future-world context.

Our contributions are threefold. 1) We introduce DriveDreamer-Policy, a unified, modular world-action architecture for autonomous driving that combines an LLM with generative experts connected through a fixed-size query interface, enabling practical compute control. 2) We incorporate an explicit 3D depth generation module and utilize a causal 3D$\rightarrow$2D$\rightarrow$1D conditioning pathway, allowing geometry to directly scaffold future video generation and motion planning and supporting multimodal data generation/simulation. (3) We conduct comprehensive experiments and analyses on Navsim~\citep{dauner2024navsimv1,cao2025navsimv2}, evaluating both planning performance and world generation quality, achieving state-of-the-art compared to existing world-based models. For instance, our model achieves an EPDMS score of 88.7 (+2.6 over the previous method) and an FVD of 53.59 (-32.36 over the previous method), improving both planning performance and video generation quality. We also perform targeted ablations to quantify how depth and future imagination jointly contribute to planning robustness.

\section{Related Works}

\subsection{Driving World-Action Models}

Driving-focused generative models use sensor data, such as images, to generate future videos, thereby enabling scalable data synthesis and simulation~\citep{wang2024drivedreamer,zhao2025drivedreamer,hassan2024gem,agarwal2025cosmos,mousakhan2025orbis,liang2025unifuture,bartoccioni2025vavimvavam,nvidia2025worldsimulationvideofoundation,ni2025recondreamer,lu2025can,zhao2025drivedreamer4d,team2025gigaworld}.
Recently, driving world action models that unify generation and planning have emerged as an active frontier. Epona~\citep{zhang2025epona} introduces an autoregressive diffusion world model that decouples causal temporal latents from per-step diffusion generation to support long-horizon video rollout and trajectory planning. ReSim~\citep{yang2025resim} trains a diffusion-transformer world simulator on real logs plus simulator non-expert behaviors to improve action-following reliability and adds Video2Reward for reward estimation. DriveVLA-W0~\citep{li2025drivevlaw0} combats the VLA supervision deficit by adding future-image world modeling and using a lightweight MoE action expert for lower latency. PWM~\citep{zhao2025pwm} treats a unified autoregressive transformer as a Policy World Model that performs action-free future forecasting and collaborative state-action prediction to benefit planning. DriveLaW~\citep{xia2025drivelaw} unifies planning and generation by feeding video-generator latents into a diffusion trajectory planner to align imagined futures with control. OmniNWM~\citep{li2025omninwm} jointly generates panoramic RGB, semantics, depth, and 3D occupancy, conditions trajectories via Plucker ray-maps, and derives intrinsic occupancy-based dense rewards. UniPGT~\citep{lu2025uniugp} unifies understanding, video generation, and trajectory planning by integrating a pretrained VLM with a video generator through hybrid experts.
Similar to existing world-action models, DriveDreamer-Policy uses an LLM to model driving-world knowledge, serving as the perception module. To incorporate multimodal generators, fixed-size latent queries are used as cross-attention keys, enabling generative experts to jointly predict depth, video, and action for the first time.

\subsection{Driving Vision-Language-Action Models}

The use of VLMs in autonomous driving has gradually evolved from scene interpretation to direct action generation. Early works such as DriveGPT4~\citep{DriveGPT4} primarily used LLMs/VLMs to generate scene descriptions and high-level maneuver suggestions, thereby improving interpretability. To bridge language understanding and low-level control, a line of modular language-to-action frameworks~\citep {zhou2025opendrivevla,CoVLA} introduced multi-stage pipelines that pass intermediate textual commands between modules; however, these non-differentiable interfaces hinder end-to-end optimization by blocking gradient backpropagation across perception, reasoning, and control. More recent end-to-end VLA methods~\citep{yang2025drivemoe,zeng2025fsdrive,team2025gigabrain} adopt unified architectures that directly map sensor observations to trajectories. Within this paradigm, DriveMoE~\citep{yang2025drivemoe} introduces a Mixture-of-Experts (MoE) design with an action decoder attached after a VLM, ReCogDrive~\citep{li2025recogdrive} couples a VLM with a diffusion planner trained by imitation and reinforcement learning to better align semantic reasoning and control, and AutoVLA~\citep{zhou2025autovla} discretizes trajectories into action primitives so that a single autoregressive model can jointly learn adaptive reasoning and planning. DriveVLA-W0~\citep{li2025drivevlaw0} scales driving VLA learning by adding dense world modeling (future image prediction) as supervision and introducing a lightweight MoE action expert to reduce inference cost and improve latency.
Our method is similar to this end-to-end VLA direction and further extends it with a unified world-action modeling design. DriveDreamer-Policy follows driving VLA models but replaces image-token world prediction with a diffusion video generation head, enabling controllable action-conditioned rollouts. We further add a depth-based 3D representation to resolve missing-geometry grounding.

\section{Methodology}

We propose DriveDreamer-Policy, a unified driving world--action model that couples a large language model with lightweight generative experts to jointly support 1) 3D world representation, 2) 2D world generation, and 3) motion planning.
We introduce world understanding with large language models and world-action prediction using generative experts in Sec.~\ref{sec:fm} and Sec.~\ref{sec:ddp}, then provide training details in Sec.~\ref{sec:impl}.


\subsection{Preliminaries}
\label{sec:fm}
\textbf{Flow Matching.}
We use conditional flow matching~\citep{lipman2023flowmatching} as the training principle for our generative experts when predicting continuous targets. 
Let $x \in \mathbb{R}^{d}$ denote a target variable, and let $c$ denote conditioning information.
Flow matching learns a time-dependent velocity field $v_{\theta}(x_t, t \,|\, c)$ that transports a simple base distribution toward the data distribution along a predefined path.
We adopt the standard linear interpolation path between a data sample $x_0 \sim p_{\text{data}}$ and a noise sample $x_1 \sim p_{\text{noise}}$:
\begin{equation}
x_t = (1-t)\,x_0 + t\,x_1, \quad t \sim \mathcal{U}(0,1),
\end{equation}
with the target velocity given by $\dot{x}_t = x_1 - x_0$.
The objective is a regression loss on the velocity:
\begin{equation}
\mathcal{L}_{\text{FM}} = \mathbb{E}_{x_0, x_1, t}\left[\left\|v_{\theta}(x_t, t \,|\, c) - (x_1-x_0)\right\|_2^2\right].
\end{equation}
At inference, we sample $x_1 \sim p_{\text{noise}}$ and integrate the induced ODE backward from $t{=}1$ to $t{=}0$ to obtain a sample consistent with the conditioning $c$.

\subsection{DriveDreamer-Policy}
\label{sec:ddp}

\begin{figure}[t]
    \centering
    \includegraphics[width=0.95\linewidth]
    {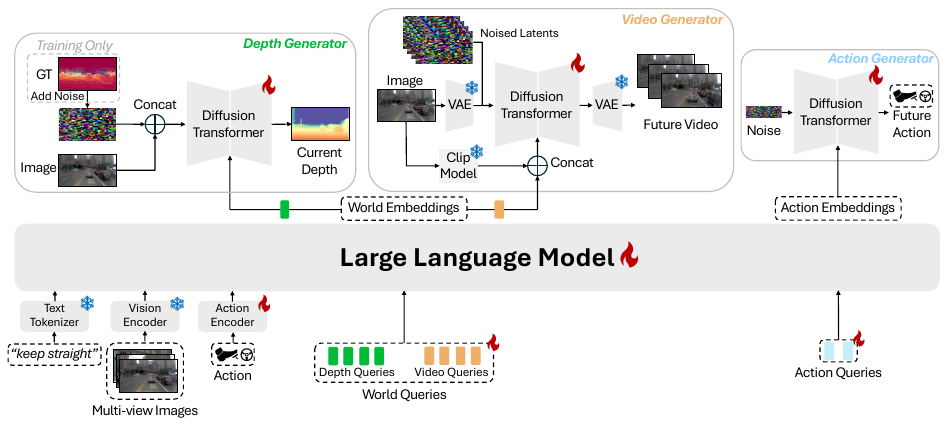}
    \captionsetup{justification=justified, singlelinecheck=false}
    \caption{Overview of our DriveDreamer-Policy pipeline. The large language model takes the language instruction, multi-view images and current action, along with a set of learnable queries as inputs to reason and generate world and action embeddings.
    The generated embeddings are then passed into our three generative expert models as cross-attention conditions to generate depth, future images, and future action.
    }
    \label{fig:pipe}
\end{figure}

Our overall pipeline is illustrated in Fig.~\ref{fig:pipe}. Multi-view images, language instructions, and the action are first encoded as tokens and processed by an LLM, together with a compact set of learned world and action queries.
The resulting world embeddings and action embeddings serve as a geometry-aware interface that conditions three modular experts: a depth generator, a video generator, and an action generator.
The LLM is responsible for multimodal understanding and producing compact state representations, while the experts generate modality-specific outputs (depth, video, and action, respectively), all mediated by a fixed-size query bottleneck.
This design is motivated by the complementary strengths of the two components: the LLM provides stable semantics and strong contextual reasoning, whereas generative experts better capture multi-modality and uncertainty in long-horizon prediction.
As a result, the model can operate in multiple modes: planning-only (enabling only the action expert), imagination-enabled planning (running action plus depth/video generation when needed), or full generation for offline simulation and data synthesis.

\subsubsection{World Understanding}
\label{sec:world_understanding}


\textbf{Input Processing.}
At each decision step, the model takes as input a natural-language instruction and synchronized multi-view RGB observations. We also provide the current action as context to the LLM, thereby contributing to both world modeling and planning.
We tokenize the inputs into three streams.
First, the instruction is converted into standard text tokens using the LLM's tokenizer.
Second, each camera view is encoded by the vision encoder into a sequence of visual patch tokens.
Third, the action context is embedded into a set of action tokens using a lightweight action encoder.
Finally, we append three fixed-size groups of learnable query tokens—depth queries, video queries, and action queries—in that order.
This design yields a stable, compact interface: the backbone always consumes the same set of query slots, and downstream heads can read out the corresponding query embeddings to produce depth maps, future videos, and future actions.

\textbf{Embeddings Generation.}
All tokens are processed by the LLM to produce contextualized hidden states. We impose a causal ordering across query groups: depth queries come first, video queries can attend to depth context, and action queries can attend to both depth and video context.
Concretely, within the same time step the attention pattern satisfies
depth queries$ \rightarrow$video queries $\rightarrow$ action queries.
This structured mask provides a clean, single-pass information flow without extra synchronization or iterative refinement across branches.


\subsubsection{World and Action Prediction}
\label{sec:world_prediction}

\textbf{Depth Generator.}
We generate a monocular depth map as an explicit 3D scaffold for the world action model.
Depth is not only a geometric output for visualization: it provides a compact representation that is directly useful for downstream video imagination (\textit{e.g.}, occlusions and object boundaries) and action planning (\textit{e.g.}, free space and distance-to-collision cues).
Our model predicts depth with a generative objective rather than a purely deterministic regression head, thereby better capturing the inherent ambiguity of monocular depth and preserving sharp depth discontinuities.
Generating depth directly in pixel-space is practical here because depth has much lower dimensionality than RGB video and preserves boundary fidelity without requiring an additional learned codec.

As shown in the upper-left of Fig.~\ref {fig:pipe}, our depth generator is a pixel-space diffusion transformer trained with a standard flow-matching objective.
During training, we sample a continuous flow time and corrupt the ground-truth depth with noise; the denoiser takes as input the concatenation of the noisy depth and the corresponding RGB image, and predicts the denoising update. To ground pixel-space generation in global scene semantics, we condition the depth denoiser on the LLM world depth embeddings through cross-attention: the depth-query world embedding acts as a compact global representation (keys/values) that guides the diffusion transformer to maintain global structural consistency while recovering fine-grained geometric details. This makes depth a queryable modality in DriveDreamer-Policy: the predicted depth can be generated on demand with depth embeddings,
and the depth embeddings serve as an upstream geometric feature that later query groups (video/action) can attend to.

\textbf{Video Generator.}
For future video generation, we employ a text-image-to-video diffusion transformer
~\citep{peebles2023dit,wan2025wan} (see the upper-middle of Fig.~\ref {fig:pipe}).
Given the current RGB images, we first encode them into a compact latent representation using a VAE and initialize a sequence of noisy video latents for the target horizon.
Instead of conditioning the diffusion model on text embedding (as in standard text-to-video pipelines), we condition it on the LLM world video embeddings produced by the video query tokens.
These world embedding tokens summarize language intent, multi-view perception, and action context, and extrinsically incorporate upstream geometric cues from the depth queries. The video denoiser attends to the world embeddings at each transformer block via cross-attention. 
To preserve appearance, identity, and camera content, we additionally extract a lightweight visual condition from the current image frame using a CLIP~\citep{radford2021clip} model and inject it into the denoiser as an explicit conditioning signal, concatenated with the world video embeddings as shown in our pipeline.
This design keeps the generator tightly grounded in the current scene and commanded maneuver, while enabling controllable, action-aware video generation.

\textbf{Action Generator.}
As presented in the upper-right of Fig.~\ref{fig:pipe}, the action generator is implemented as a standalone diffusion transformer that maps a noise trajectory to a feasible future action sequence.
It is conditioned on the action embedding produced by the LLM from the action query tokens, which aggregates instruction semantics, multi-view observations, and upstream geometric and imagination cues.
This conditioning is injected via cross-attention, thereby keeping the action head lightweight while still leveraging rich scene context.
Since the action generator does not depend on explicit depth and video generation, it can be activated independently for planning, while implicitly benefiting from predicted future world context.

We parameterize each trajectory state by position and heading using a continuous representation~\citep{zhou2024smartrefine}, namely $(x, y, \cos\theta, \sin\theta)$, which avoids angular wrap-around and encourages smooth turn dynamics.


\subsection{Training Details}
\label{sec:impl}

\textbf{{Depth Normalization.}}
We normalize depth to a stable range before training the depth generator.
Given a depth map, we first apply a log transform and then compute per-map percentiles to normalize to the range [-0.5, 0.5]. During inference, we invert the transform to recover metric or relative depth as needed.

\textbf{{Model Initialization and Adaptation.}}
For the large language model, we use Qwen3-VL-2B~\citep{bai2025qwen3vl} to process and understand multimodal inputs. For the depth generator, we initialize our model from PPD~\citep{xu2025pixelperfectdepth}. For the video generator, we initialize the model from Wan-2.1-T2V-1.3B~\citep{wan2025wan} and adapt it to the image-to-video task. For both depth and video generation, we fine-tune at a spatial resolution of $144 \times 256$ to reduce computational and memory costs. The video training horizon is 9 frames.

\textbf{{Training Objective and Optimization.}}
We train all components in a single stage with a joint multi-task loss:
\begin{equation}
\mathcal{L} = \lambda_{d}\mathcal{L}_{d} + \lambda_{v}\mathcal{L}_{v} + \lambda_{a}\mathcal{L}_{a},
\end{equation}
where $\mathcal{L}_{d}$ is the loss for depth prediction, $\mathcal{L}_{v}$ is the loss for video prediction, and $\mathcal{L}_{a}$ is the loss for trajectory prediction. We use $\lambda_{d}=0.1$ and set the remaining hyperparameters to 1.0 by default. The depth label used in training is obtained from an off-the-shelf depth foundation model, Depth Anything 3 (DA3)~\citep{lin2025depth3}.
\section{Experiments}

\subsection{Experimental Setup}

\textbf{Datasets and Planning Metrics.}
We train and evaluate our method on the Navsim benchmark~\citep{dauner2024navsimv1, cao2025navsimv2}, which is derived from real-world driving logs and provides synchronized surround-view sensory inputs for end-to-end planning evaluation. Following the standard Navsim protocol, we train on the \texttt{navtrain} split and evaluate on the \texttt{navtest} split, which contains 100k and 12k data samples sampled at 2Hz, respectively.
Navsim evaluates closed-loop planning performance using the predictive driver model score (PDMS) on v1 and the extended PDMS (EPDMS) on v2.
PDMS aggregates multiple safety and quality terms, including no-at-fault collision, drivable-area compliance, time-to-collision, ego progress, and comfort; EPDMS further includes direction and traffic-light compliance, as well as lane-keeping and comfort.
For Navsim-v2, we align with the common practice in recent methods~\citep{Liao_2025_CVPR,li2025drivevlaw0} and evaluate EPDMS on the \texttt{navtest} split for fair and convenient comparison.

\textbf{World Generation Metrics.}
In addition to planning, we evaluate our generative experts.
Video is evaluated on Navim using the recorded future RGB frames as ground truth. Depth is evaluated using dense depth targets provided by DA3, which are also used for training. 
We report Absolute Relative Error (AbsRel) to quantify relative
depth differences and threshold accuracy ($\delta$) to measure the
proportion of accurate predictions within a specified relative
error. Higher $\delta$ values and lower AbsRel values indicate
better depth estimation performance.
For video evaluation, we follow~\citep{zhao2025pwm} and report perceptual quality and temporal consistency of predicted future frames using Learned Perceptual Image Patch Similarity (LPIPS)~\citep{Zhang_2018_CVPR}, Peak Signal-to-Noise Ratio (PSNR)~\citep{Huynh-Thu2008-qw} and Fréchet Video Distance (FVD)~\citep{unterthiner2019fvd}.

\begin{table}[t]
\centering
\captionsetup{justification=justified, singlelinecheck=false}
\caption{Comparison with state-of-the-art methods on the the \texttt{navtest} of Navsim v1~\citep{dauner2024navsimv1} benchmark. The best result outlines as bold. ``*'' indicates results with imitation learning. Methods fall into three categories: vision-based E2E models, VLA-based models, and world-model-based models.
}
\renewcommand\arraystretch{0.8}
\resizebox{0.95\linewidth}{!}{
\begin{tabular}{ll|c|ccccc|c}
\toprule
Methods & Venue & Sensors & NC$\uparrow$ & DAC$\uparrow$ & TTC$\uparrow$ & C$\uparrow$ & EP$\uparrow$ & PDMS$\uparrow$ \\
\midrule
Human    &   ~~~~--  & ~~~~-- & 100.0 &  100.0   &  100.0   &  99.9 &  87.5  &   94.8   \\
\midrule
\rowcolor{lightgrey} \multicolumn{9}{c}{\textit{Vision-Based End-to-End Methods}}\\
TransFuser~\citep{chitta2022transfuser}   & TPAMI'23 &  3$\times$C ${+}$ L   &  97.7  &  92.8   &   92.8  &  \textbf{100.0} &  79.2  &  84.0  \\
UniAD~\citep{hu2023uniad}   & CVPR'23 &  6$\times$C   &  97.8  &  91.9   &   92.9  &  \textbf{100.0} &  78.8  &  83.4  \\
PARA-Drive~\citep{10656117}    & CVPR'24  &  6$\times$C   &  97.9  &  92.4   &  93.0  &  99.8 &  79.3  &  84.0  \\
DiffusionDrive~\citep{liao2025diffusiondrive}  &  CVPR'25 & 3$\times$C ${+}$ L &  98.2  &  96.2   &  94.7   & \textbf{100.0}  &  82.2  &   88.1   \\
\midrule
\rowcolor{lightgrey} \multicolumn{9}{c}{\textit{Vision-Language-Action Methods}}\\
AutoVLA~\citep{zhou2025autovla}   & NeurIPS'25 &  3$\times$C   &  98.4  &  95.6   &   \textbf{98.0}  &  99.9 &  81.9  &  89.1  \\
Recogdrive$^{*}$~\citep{li2025recogdrive}   & ICLR'26 &  3$\times$C   &  98.1  &  94.7   &   94.2  &  \textbf{100.0} &  80.9  &  86.5  \\
DriveVLA-W0~\citep{li2025drivevlaw0}   & ICLR'26 &  1$\times$C   &  98.7  &  96.2   &   95.5  &  \textbf{100.0} &  82.2  &  88.4  \\

\midrule
\rowcolor{lightgrey} \multicolumn{9}{c}{\textit{World-Model-Based Methods}}\\
LAW~\citep{li2025law}   & ICLR'25 &  1$\times$C   &  96.4  &  95.4   &   88.7  &  99.9 &  81.7  &  84.6  \\
DrivingGPT~\citep{Chen_2025_ICCV}&ICCV'25&1$\times$C&\textbf{98.9}& 90.7& 94.9& 95.6& 79.7& 82.4\\
WoTE~\citep{Li_2025_ICCV} & ICCV'25 &  3$\times$C ${+}$ L   &  98.5  &  96.8   &   94.4  &  99.9 &  81.9  &  88.3  \\
Epona~\citep{zhang2025epona}&ICCV'25&3$\times$C&97.9& 95.1& 93.8& 99.9& 80.4& 86.2\\
FSDrive~\citep{zeng2025fsdrive} & NeurIPS'25 & 3$\times$C & 98.2 & 93.8 & 93.3 & 99.9 & 80.1 & 85.1\\
PWM~\citep{zhao2025pwm} & NeurIPS'25 & 1$\times$C & 98.6 & 95.9 & 95.4 & \textbf{100.0} & 81.8 & 88.1\\
\rowcolor{cyan!15}DriveDreamer-Policy (Ours) & ~~~~-- & 3$\times$C & 98.4 & \textbf{97.1} & 95.1 & \textbf{100.0} & \textbf{83.5} & \textbf{89.2}\\
\bottomrule
\end{tabular}
}
\label{tab:navsimv1}
\end{table}

\begin{table}[!t]
\centering
\captionsetup{justification=justified, singlelinecheck=false}
\caption{Comparison with state-of-the-art methods on the \texttt{navtest} of Navsim v2~\citep{cao2025navsimv2}.
}
\renewcommand\arraystretch{0.8}
\resizebox{.95\linewidth}{!}{
\begin{tabular}{ll|ccccccccc|c}
\toprule
Methods & Venue & NC$\uparrow$ & DAC$\uparrow$ & DDC$\uparrow$ & TLC$\uparrow$ & EP$\uparrow$ & TTC$\uparrow$ & LK$\uparrow$ & HC$\uparrow$ & EC$\uparrow$ & EPDMS$\uparrow$ \\
\midrule
\midrule
\rowcolor{lightgrey} \multicolumn{12}{c}{\textit{Vision-Based End-to-End Methods}} \\
TransFuser~\citep{chitta2022transfuser} & TPAMI'23& 96.9 & 89.9 & 97.8 & 99.7 & 87.1 & 95.4 & 92.7 & \textbf{98.3} & 87.2 & 76.7 \\
DiffusionDrive~\citep{liao2025diffusiondrive} & CVPR'25 & 98.2 &  95.9 & 99.4 & 99.8 & 87.5 & 97.3 & 96.8 & \textbf{98.3} & 87.7 & 84.5\\
Drivesuprim~\citep{yao2025drivesuprim} & AAAI'26 &97.5& 96.5& 99.4& 99.6& \textbf{88.4}& 96.6& 95.5& \textbf{98.3}& 77.0& 83.1\\
ARTEMIS~\citep{11248836} & RAL'26 &98.3& 95.1& 98.6& 99.8 & 81.5& 97.4& 96.5& \textbf{98.3}& \textbf{89.1} & 83.1\\
\midrule
\rowcolor{lightgrey} \multicolumn{12}{c}{\textit{Vision-Language-Action Methods}} \\
DriveVLA-W0~\citep{li2025drivevlaw0} & ICLR'26 &\textbf{98.5} & \textbf{99.1}& 98.0& 99.7& 86.4& \textbf{98.1}& 93.2& 97.9& 58.9& 86.1 \\
\midrule
\rowcolor{lightgrey} \multicolumn{12}{c}{\textit{World-Model-Based Methods}}\\
\rowcolor{cyan!15}DriveDreamer-Policy (Ours)& ~~~~~-- & 98.4 & 97.1 & \textbf{99.5} & \textbf{99.9} & 87.9 & 97.7 & \textbf{97.6} & \textbf{98.3} & 79.4 & \textbf{88.7} \\
\bottomrule
\end{tabular}
}
\label{tab:navsimv2}
\end{table}

\begin{table}[!t]
\centering
\captionsetup{justification=justified, singlelinecheck=false}
\caption{World generation performance on Navsim. 
For video generation, we compare against existing generative world-model methods trained and evaluated on Navsim. 
For depth generation, we report comparisons among our model variants. Our method achieves higher quality for depth and video.}
\renewcommand\arraystretch{0.8}
\resizebox{\linewidth}{!}{
\begin{minipage}{0.52\linewidth}
    \centering
    \small (a) Video performance comparison.
    \resizebox{\linewidth}{!}{
    \begin{tabular}{ll|cc|c}
    \toprule
    Methods & Venue & LPIPS$\downarrow$ & PSNR$\uparrow$ & FVD$\downarrow$ \\
    \midrule
    PWM~\citep{zhao2025pwm} & NeurIPS'25 & 0.23 & \textbf{21.57} & 85.95 \\ 
    \rowcolor{cyan!15}DriveDreamer-Policy (Ours) & ~~~~-- & \textbf{0.20} & 21.05 & \textbf{53.59}\\
    \bottomrule
    \end{tabular}
    }
    \label{tab:exp_video_left}
\end{minipage}
\hfill
\begin{minipage}{0.48\linewidth}
    \centering
    \small (b) Depth performance comparison.
    \resizebox{.9\linewidth}{!}{
    \begin{tabular}{ll|c|ccc}
    \toprule
    Methods & Venue & AbsRel$\downarrow$ & $\delta_1$$\uparrow$ & $\delta_2$$\uparrow$ & $\delta_3$$\uparrow$ \\
    \midrule
    PPD~\citep{xu2025pixelperfectdepth} &NeurIPS'25  & 18.5 & 80.4 & 94.0 & 97.2 \\
    PPD-Fintuned & NeurIPS'25  & 9.3& 91.4& 98.3& \textbf{99.5} \\
    \rowcolor{cyan!15}DriveDreamer-Policy (Ours) & ~~~~-- & \textbf{8.1} & \textbf{92.8} & \textbf{98.6} & \textbf{99.5} \\
    \bottomrule
    \end{tabular}
    }
    \label{tab:exp_video_right}
\end{minipage}
}
\label{res:world}
\end{table}

\textbf{Baselines.}
We compare against strong Navsim baselines spanning three families.
1) Classical vision-based end-to-end planners that utilize vision models and map sensor inputs to trajectories, including TransFuser~\citep{chitta2022transfuser}, UniAD~\citep{hu2023uniad} and DiffusionDrive~\citep{liao2025diffusiondrive}.
2) Vision-language-action planners that use large language models and predict trajectories as tokens or by a diffusion-based expert, including approaches: DriveVLA-W0~\citep{li2025drivevlaw0}, AutoVLA~\citep{zhou2025autovla} and Recogdrive~\citep{li2025recogdrive}.
3) World-model-based planners that integrate foresight into planning, including  LaW~\citep{li2025law}, DrivingGPT~\citep{DriveGPT4}, WoTE~\citep{Li_2025_ICCV}, Epona~\citep{zhang2025epona}, FSDrive~\citep{zeng2025fsdrive} and PWM~\citep{zhao2025pwm}.
All baselines are reported under their official Navsim performance.

\textbf{Implementation Details.}
We implement the action encoder as a 2-layer MLP with layer normalization~\citep{ba2016layernormalization}.
We train DriveDreamer-Policy in a single stage for 100k steps with a batch size of 32 on 8 NVIDIA H20 GPUs, using AdamW optimizer~\citep{loshchilov2019adamw} with a learning rate of $1\times10^{-5}$. 
Unless otherwise stated, all experiments use the same query configuration (64 depth-query tokens, 64 video-query tokens, and 8 action-query tokens). 
We use Navsim training data, without additional datasets or extra pre-training beyond the initialized backbones.

\subsection{Quantitative Results}

\begin{table}[!t]

\def\arraystretch{0.8}
    \centering
    \captionsetup{justification=justified, singlelinecheck=false}
    \caption{Ablations on World Learning for Planning. All world learning strategies effectively improve planning performance compared to training from scratch.
    }
    \resizebox{0.95\linewidth}{!}{
    \begin{tabular}{c|cc|ccccc|c}
        \toprule
             {\multirow{1}{*}{\makecell[c]{Strategy }}} 
             & \multicolumn{2}{c|}{\makecell[c]{World Representation}}
             &\multicolumn{6}{c}{\makecell[c]{Planning Metrics}} \\
             \cmidrule(r){2-9}
               & \makecell[c]{~~~Depth} & \makecell[c]{Video}  & NC$\uparrow$ & DAC$\uparrow$ & TTC$\uparrow$ & C$\uparrow$ & EP$\uparrow$ & PDMS$\uparrow$   \\ 
            \midrule
            \multirow{1}{*}{Without World Learning}
            & \mineno & \mineno & 98.0 & 96.3 & 94.4 & \textbf{100.0} & 82.5& 88.0\\
            \midrule
            \multirow{3}{*}{\shortstack{With World Learning}}
            & \mineyes  &  \mineno
            & 98.1 & 96.7 & 94.9 & \textbf{100.0} & 82.8 & 88.5 \\
            & \mineno  &  \mineyes
            & 98.1 & 97.0 & 95.0 & \textbf{100.0} & 83.1 & 88.9 \\
            & \mineyes  &  \mineyes
            & \textbf{98.4} & \textbf{97.1} & \textbf{95.1} & \textbf{100.0} & \textbf{83.5} & \textbf{89.2}    \\
        \bottomrule
    \end{tabular}}
    \label{tab:abi_world}
\end{table}

\begin{table}[!t]

\def\arraystretch{0.8}
    \centering
    \captionsetup{justification=justified, singlelinecheck=false}
    \caption{Ablations on Depth Learning for Video Generation. Using depth as a prior in joint learning improves video generation accuracy.
    }
    \resizebox{0.75\linewidth}{!}{
    \begin{tabular}{c|c|ccc}
        \toprule
             {\multirow{1}{*}{\makecell[c]{Strategy }}} 
             & \multicolumn{1}{c|}{\makecell[c]{World Representation}}
             &\multicolumn{3}{c}{\makecell[c]{Video Metrics}} \\
             \cmidrule(r){2-5}
               & \makecell[c]{Depth} & LPIPS$\downarrow$ & PSNR$\uparrow$ & FVD$\downarrow$   \\ 
            \midrule
            \multirow{1}{*}{Without Depth Learning}
            & \mineno & 0.22 & 19.89 & 65.82 \\
            \midrule
            \multirow{1}{*}{\shortstack{With Depth Learning}}
            & \mineyes & \textbf{0.20} & \textbf{21.05} & \textbf{53.59} \\
        \bottomrule
    \end{tabular}}
    \label{tab:abi_video}
\end{table}

\begin{table}[!t]

\def\arraystretch{0.8}
    \centering
    \captionsetup{justification=justified, singlelinecheck=false}
    \caption{Ablations on Number of Queries. More query tokens provide higher-capacity slots for storing relevant context, thereby enhancing both generation and planning.
    }
    \resizebox{0.99\linewidth}{!}{
    \begin{tabular}{ccc|cccc|ccc|ccccc|c}
        \toprule
        \multicolumn{3}{c|}{\makecell[c]{Number of Queries}}
        & \multicolumn{13}{c}{\makecell[c]{Metrics}} \\
        \midrule
        \multirow{2}{*}{\makecell[c]{Depth}}
        & \multirow{2}{*}{\makecell[c]{Video}}
        & \multirow{2}{*}{\makecell[c]{Action}}
        & \multicolumn{4}{c|}{Depth}
        & \multicolumn{3}{c|}{Video}
        & \multicolumn{6}{c}{Action} \\
        \cmidrule(lr){4-7}\cmidrule(lr){8-10}\cmidrule(lr){11-16}
        & & & AbsRel$\downarrow$ & $\delta_1$$\uparrow$ & $\delta_2$$\uparrow$ & $\delta_3$$\uparrow$
            & LPIPS$\downarrow$ & PSNR$\uparrow$ & FVD$\downarrow$
            & NC$\uparrow$ & DAC$\uparrow$ & TTC$\uparrow$ & C$\uparrow$ & EP$\uparrow$ & PDMS$\uparrow$ \\
        \midrule
        32 & 32 & 4 &  9.7 & 90.2 & 97.9 & 99.4 & \textbf{0.20} & 20.67 & 57.97& 98.2& 97.0&95.0 & \textbf{100.0} & 83.2 & 88.9\\
        64 & 64 & 8 & \textbf{8.1} & \textbf{92.8} & \textbf{98.6} & \textbf{99.5} & \textbf{0.20} & \textbf{21.05} & \textbf{53.59} & \textbf{98.4} & \textbf{97.1} & \textbf{95.1} & \textbf{100.0} & \textbf{83.5} & \textbf{89.2} \\
        \bottomrule
    \end{tabular}}
    \label{tab:abi_query}
\end{table}

\textbf{Planning Performance Comparison with Other Methods.}
As shown in Table~\ref{tab:navsimv1} and Table~\ref{tab:navsimv2},
we first report planning performance compared with other methods, on the \texttt{navtest} set of Navsim v1 using PDMS and Navsim v2 using EPDMS.
For a fair and meaningful comparison, we consider only methods that were published at the time of this paper's submission.
Specifically, we categorize comparisons against state-of-the-art vision-based end-to-end planners, vision-language-action planners, and world-model-based models. Our DriveDreamer-Policy outperforms all considered methods and achieves PDMS scores of 89.2 and 88.7 on Navsim v1 and v2, respectively. Moreover, DriveDreamer-Policy achieves strong performance on specific planning-critic sub-scores, indicating our method's control of planning trajectories (\textit{e.g.}, a DAC of 97.1 and an EP of 83.5 on Navsim v1, and a DDC of 99.5 and an LK of 97.6 on Navsim v2).

\textbf{World Performance Comparison with Other Methods.}
We also compare with other generative world-based methods. For video quality, we compare with the existing method PWM~\citep{zhao2025pwm}. Since PWM supports only single-view generation, we evaluate the single-view (front) quality for a fair and convenient comparison. As presented in Table~\ref{tab:exp_video_left}(a), our method shows a larger improvement than PWM by a substantial margin (\textit{e.g.}, improvement of 32.36 on FVD).
For depth accuracy, we compare against PPD, because our depth generator is initialized from it. To the best of our knowledge, there is no widely adopted Navsim benchmark that reports directly comparable results for depth prediction.
Specifically, we evaluate two variants: (i) zero-shot PPD on Navsim and (ii) fine-tune PPD on Navsim.
As shown in Table~\ref{tab:exp_video_right}(b), DriveDreamer-Policy achieves lower depth error. We attribute the improvement to LLM conditioning: the depth denoiser is guided by world depth embeddings, resolving locally ambiguous regions and producing more consistent geometry than image cues alone.

\begin{figure}[t]
    \centering
    \includegraphics[width=0.96\linewidth]{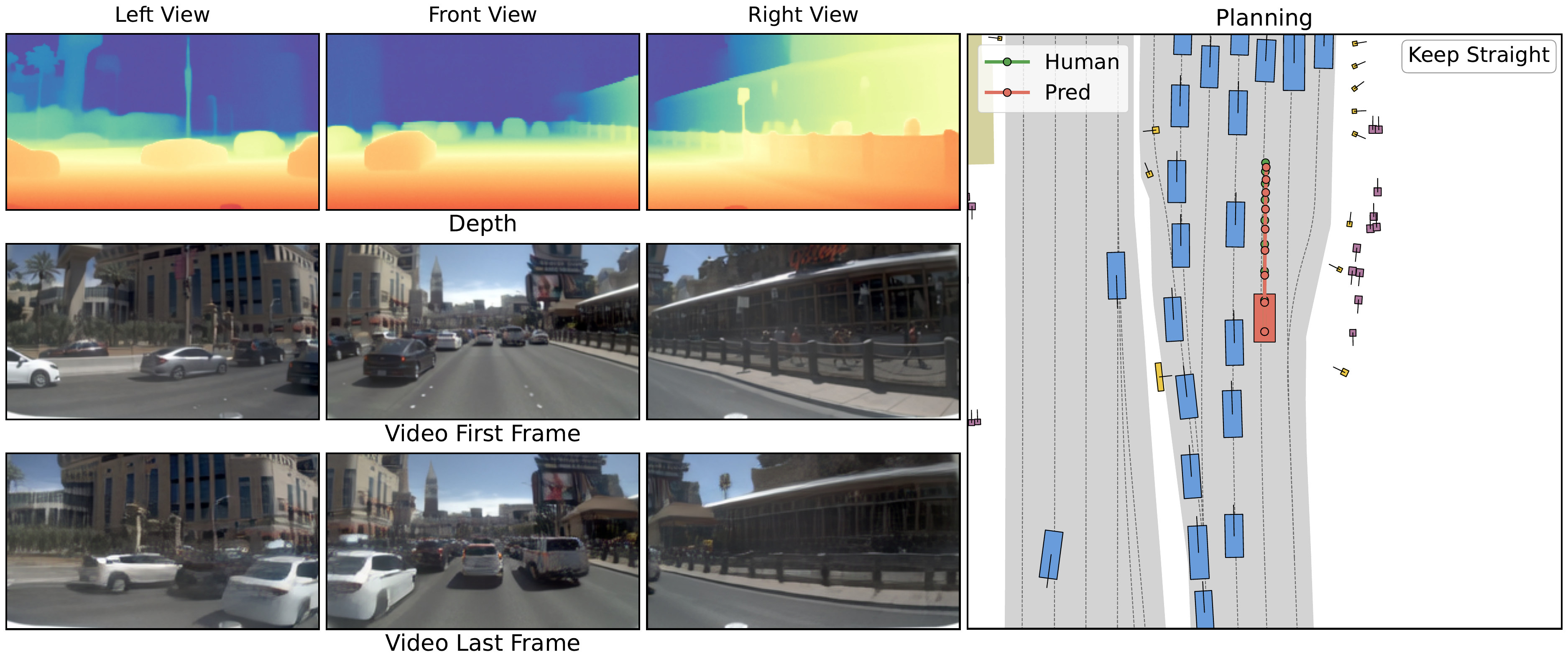}
    
    \includegraphics[width=0.96\linewidth]{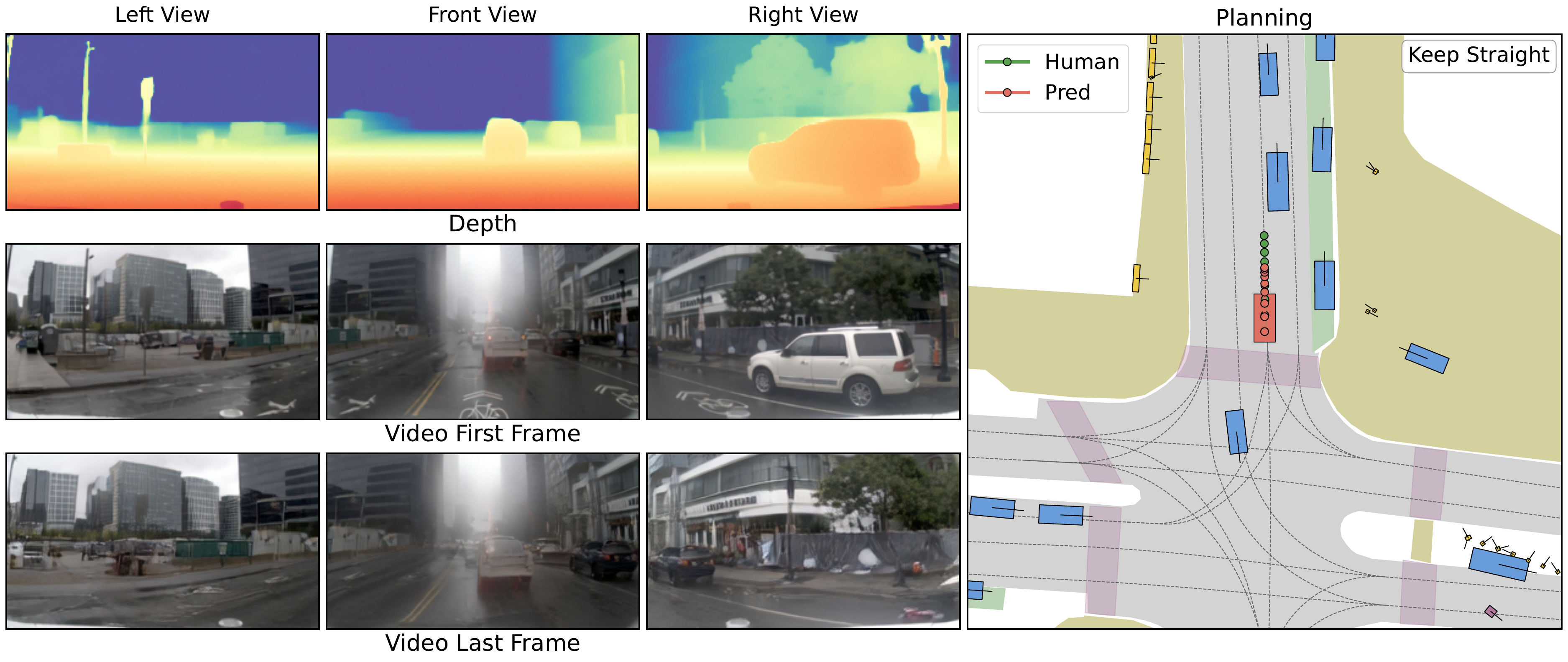}

    \captionsetup{justification=justified, singlelinecheck=false}
    \caption{Visualization Results of our method. 
    We show the generated depth, video, and actions, respectively. Depth is truncated to below 80 meters for better visualization. Our generation results remain spatially stable, and the planning performs well compared with human trajectories (\textit{e.g.}, aligns with human trajectories (top) and slows down more effectively than human trajectories (bottom)).
}
    \label{fig:vis_gen_plan}
\end{figure}

\begin{figure}[!t]
    \centering
    
    \includegraphics[width=0.95\linewidth]{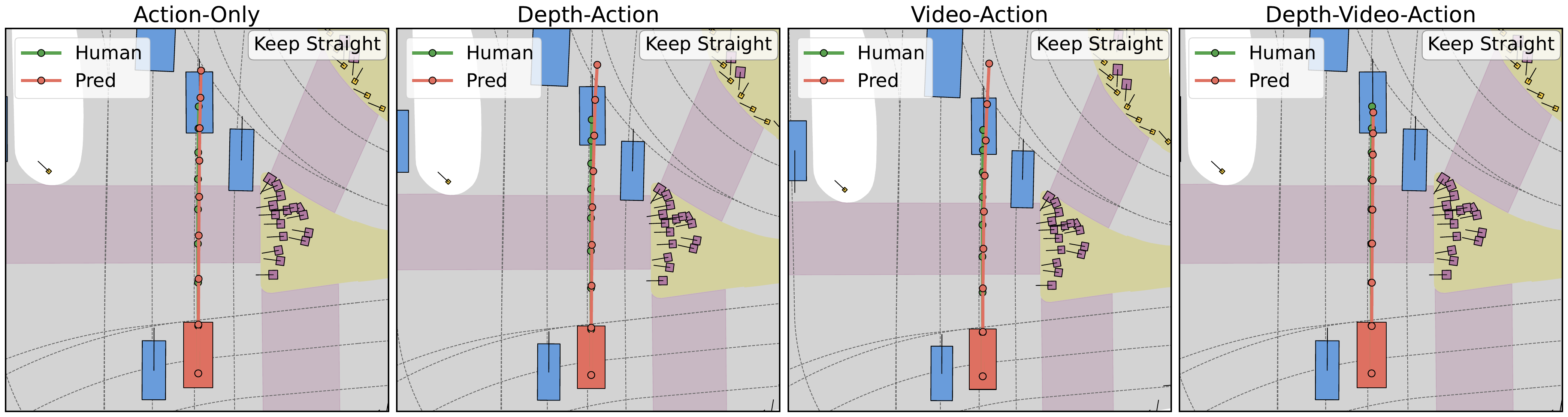}

    \includegraphics[width=0.95\linewidth]{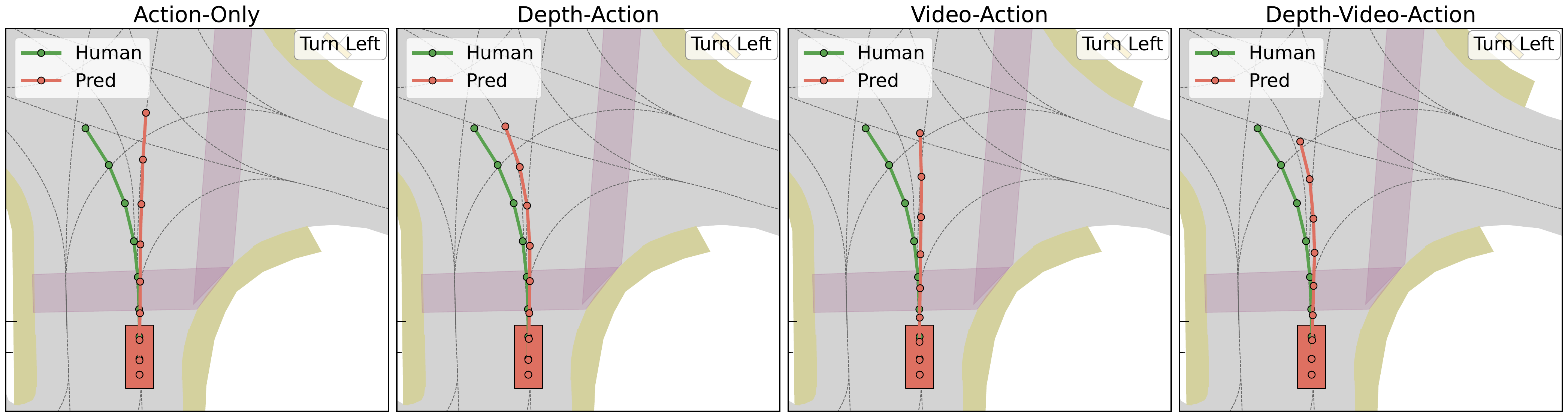}
    
    \includegraphics[width=0.95\linewidth]{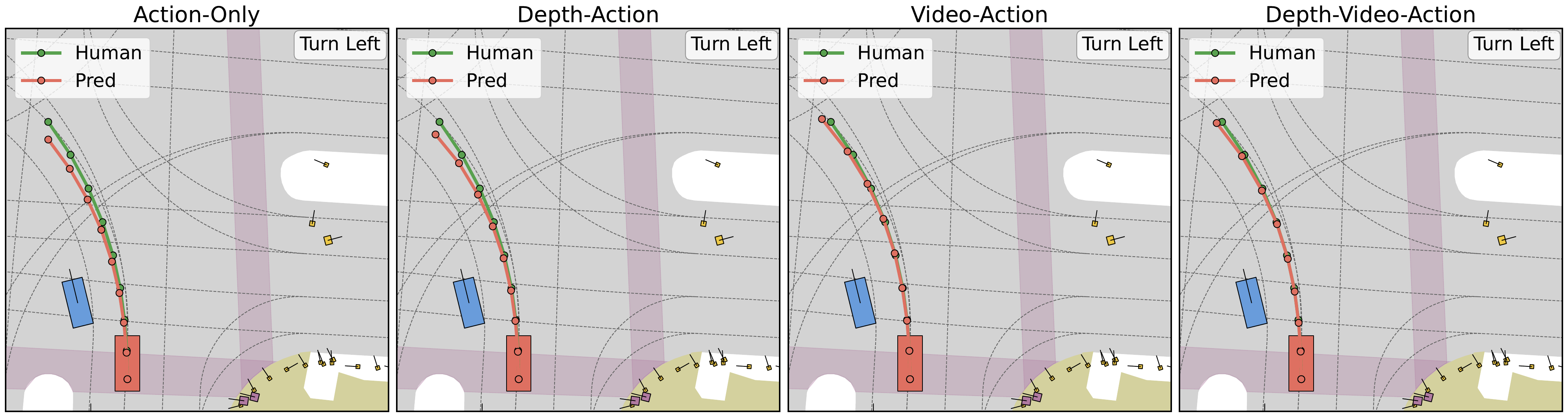}

    \captionsetup{justification=justified, singlelinecheck=false}
    \caption{Visualization of world learning for planning.
    Columns compare Action-Only, Depth-Action, Video-Action, and Depth-Video-Action variants.
    Green denotes the human (expert) trajectory and red denotes the predicted trajectory.
    The three rows correspond to (top) avoiding potential collision by a slower trajectory, (middle) correcting an initially wrong maneuver, and (bottom) aligning more closely with the human trajectory.
    Depth and video provide complementary world cues that improve safety margins and trajectory consistency.
}
    \label{fig:vis_world_4_plan}
\end{figure}

\textbf{Ablations on World Learning for Planning.}
To isolate the contribution of each modality, we evaluate four variants under identical training budgets:
1) action-only: the simplest VLA version without any world modeling,
2) depth+action: generates depth and future action,
3) video+action: generates video and future action,
and 4) depth+video+action: our full model.
We report improvements in PDMS for each variant. The results are shown in Table~\ref{tab:abi_world}. All world training strategies effectively improve planning performance compared to training from scratch. However, single-world training (depth or video) yields smaller improvement, likely due to incomplete world information. Joint world training, leveraging both depth geometry and video temporal evolution, achieves the largest performance gains by enabling the model to learn more generalizable and robust features.

\textbf{Ablations on Depth Learning for Video Generation.}
The impact of depth learning on future video generation is evaluated by comparing two variants under the same training data and compute budget: 1) video-only, where the video generator is conditioned on the backbone features without depth joint learning; 2) depth+video, where depth is trained jointly and the video queries are causally conditioned on depth queries.
Video quality is reported in Table~\ref{tab:abi_video}. Joint learning with depth improves video generation accuracy, indicating that depth provides an effective 3D scaffold for coherent future prediction.

\subsection{Ablation Studies}

\textbf{Ablations on Number of Queries.}
We ablate the query budget to study the trade-off between accuracy. Specifically, we compare the default setting (64 depth + 64 video + 8 action query tokens) against smaller budgets (32 depth + 32 video + 4 action query tokens).
The results, presented in Table~\ref{tab:abi_query}, indicate that increasing the query budget generally improves both planning and world-generation performance, as more query tokens provide higher-capacity slots to store geometry, appearance, and action-relevant context.

\subsection{Qualitative Results}

\textbf{Visualizations of World generation and Motion Planning.}
We visualize representative Navsim scenarios in Fig.~\ref{fig:vis_gen_plan} by overlaying predicted trajectories on BEV renderings and by showing the corresponding generated depth maps and future frames.
These examples highlight how depth-conditioned imagination reduces common failure modes (\textit{e.g.}, short-horizon collision risk, off-road drift) and improves interpretability by exposing the model's predicted geometry and future scene evolution.

\textbf{Visualizations of World Learning for Planning.} We present additional visualization results to demonstrate the effectiveness of world learning in benefiting planning.
Fig.~\ref{fig:vis_world_4_plan} presents qualitative comparisons that demonstrate how learned world representations improve planning.
Each row visualizes a representative Navsim scenario, and each column corresponds to an ablation variant: Action-Only, Depth-Action, Video-Action, and Depth-Video-Action.
Across the three cases, adding world learning consistently produces safer and more human-like trajectories.
In the top-row example, action-only planning tends to drift toward conflicting traffic, whereas depth- and video-conditioned variants maintain a clearer safety margin.
In the middle row example, world learning helps recover the correct maneuver early, reducing lateral deviation and preventing late, abrupt turns.
In the bottom-row example, the full model yields trajectories that more closely match the expert path, suggesting that geometry cues (depth) and future appearance/dynamics cues (video) provide complementary guidance for action prediction.

\section{Conclusion}
We present DriveDreamer-Policy, a unified driving world–action model that jointly performs depth generation, future video imagination, and motion planning within a single framework. The model combines a large language model with modular generative experts connected through a compact query interface, enabling flexible operating modes for both planning and world generation. By introducing depth as an explicit geometric scaffold and organizing information flow in a depth→video→action manner, the model allows planning to leverage both scene geometry and predicted future dynamics. Experiments on Navsim demonstrate strong performance across planning and world generation tasks, and ablations show that depth and video provide complementary cues that improve planning robustness.

\clearpage
\setcitestyle{numbers}
\bibliographystyle{unsrtnat}
\bibliography{main}

\end{document}